\def\year{2020}\relax
\documentclass[letterpaper]{article} 
\usepackage{aaai20}  
\usepackage{times}  
\usepackage{helvet} 
\usepackage{courier}  
\usepackage[hyphens]{url}  
\usepackage{graphicx} 
\usepackage{graphicx}  
\usepackage{amsmath}
\usepackage{amssymb}
\usepackage{amsthm}
\usepackage{bm}
\usepackage{subfigure} 
\usepackage{multirow}
\usepackage{url}
\usepackage{algorithm}
\usepackage{algorithmic}

\title{Gromov-Wasserstein Factorization Models for Graph Clustering}
\author{Hongteng Xu$^{1,2}$\\
$^1$Infinia ML Inc.\quad $^2$Department of ECE, Duke Univeristy\\
hongteng.xu@duke.edu
}
\begin{document}

\maketitle

\begin{abstract}
We propose a new nonlinear factorization model for graphs that are with topological structures, and optionally, node attributes.
This model is based on a pseudometric called Gromov-Wasserstein (GW) discrepancy, which compares graphs in a relational way.
It estimates observed graphs as GW barycenters constructed by a set of atoms with different weights. 
By minimizing the GW discrepancy between each observed graph and its GW barycenter-based estimation, we learn the atoms and their weights associated with the observed graphs.
The model achieves a novel and flexible factorization mechanism under GW discrepancy, in which both the observed graphs and the learnable atoms can be unaligned and with different sizes. 
We design an effective approximate algorithm for learning this Gromov-Wasserstein factorization (GWF) model, unrolling loopy computations as stacked modules and computing gradients with backpropagation. 
The stacked modules can be with two different architectures, which correspond to the proximal point algorithm (PPA) and Bregman alternating direction method of multipliers (BADMM), respectively. 
Experiments show that our model obtains encouraging results on clustering graphs.
\end{abstract}

\section{Introduction}
As an important methodology for machine learning, factorization models explore intrinsic structures of high-dimensional observations explicitly, which have been widely used in many learning tasks, $e.g.$, data clustering~\cite{ng2002spectral}, dimensionality reduction~\cite{candes2011robust}, recommendation systems~\cite{wang2011collaborative}, etc. 
In particular, factorization models decompose high-dimensional observations into a set of atoms under specific criteria and achieve their latent representations accordingly.
For each observation, its latent representation corresponds to the coefficients associated with the atoms.\footnote{In this paper, we borrow the terms ``atoms'' and ``coefficients'' from a kind of factorization model called dictionary learning~\cite{aharon2006k}.}

However, most of the existing factorization models, such as principal component analysis (PCA)~\cite{pearson1901liii}, nonnegative matrix factorization (NMF)~\cite{sra2006generalized}, and dictionary learning~\cite{aharon2006k}, are designed for vectorized samples with the same dimension. 
They are inapplicable to structural data, $e.g.$, graphs and point clouds. 
For example, in the task of graph clustering, the observed graphs are often with different numbers of nodes, and the correspondences between their nodes are often unknown, $i.e.$, the graphs are unaligned.
These unaligned graphs cannot be represented as vectors directly. 
Although many graph embedding methods have been proposed for these years with the help of graph neural networks~\cite{kipf2016semi,ying2018hierarchical}, they often require side information like node attributes and labels, which may not be available in practice.
Moreover, without explicit factorization mechanisms, these methods cannot find the atoms that can reconstruct observed graphs, and thus, the graph embeddings derived by them are not so interpretable as the latent representation derived by factorization models. 
Therefore, it is urgent to build a flexible factorization model applicable to structural data. 

\begin{figure}[!t]
    \centering
    \includegraphics[width=0.75\linewidth]{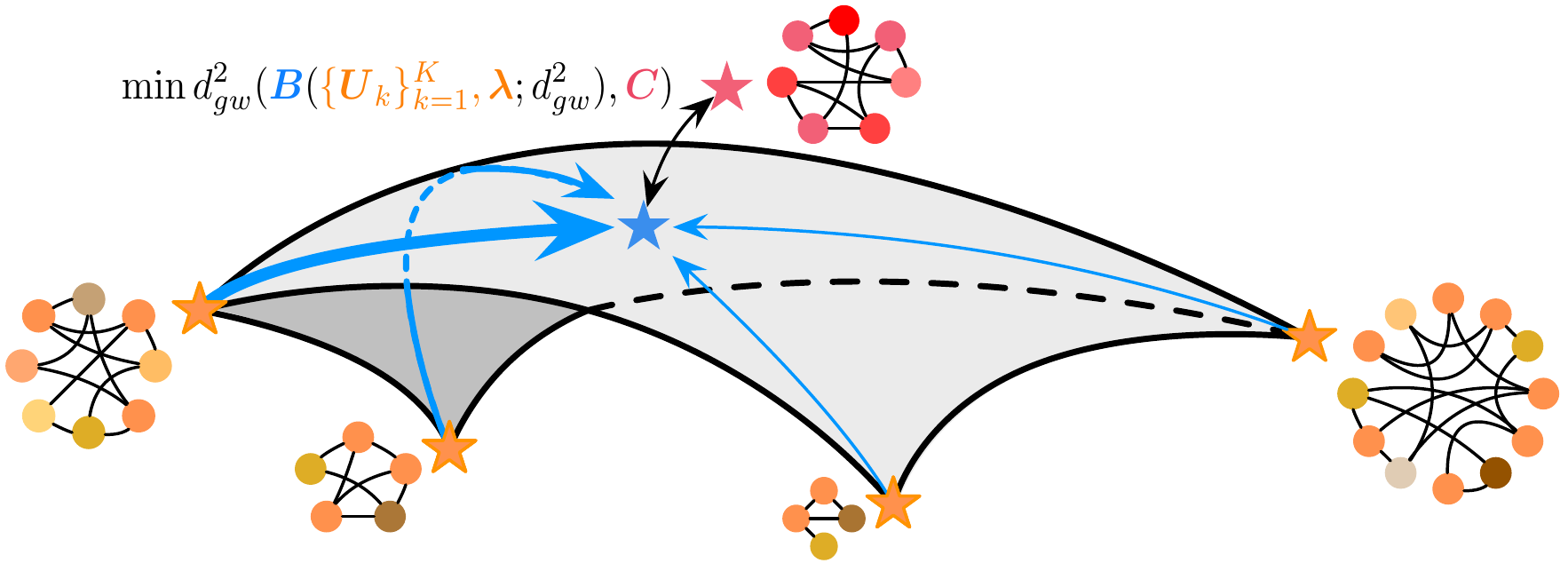}
    \caption{\small{An illustration of our Gromov-Wasserstein factorization model. Each star indicates a graph. For each graph, the black curves show its edges, and the dots with different colors are its nodes with different attributes.}}
    \label{fig:scheme}
\end{figure}

To overcome the challenges above, we propose a novel Gromov-Wasserstein factorization (GWF) model based on Gromov-Wasserstein (GW) discrepancy~\cite{memoli2011gromov,chowdhury2018gromov} and barycenters~\cite{peyre2016gromov}.  
As illustrated in Fig.~\ref{fig:scheme}, for each observed graph ($i.e.$, the red star), our GWF model reconstructs it based on a set of atoms ($i.e.$, the orange stars corresponding to four graphs). 
The reconstruction ($i.e.$, the blue star) is the GW barycenter of the atoms that minimizes the GW discrepancy to the observed graph. 
The weights of the atoms ($i.e.$, the blue arrows with different widths) formulate the embedding of the observed graph. 
Learning GW barycenters reconstructs graphs from atoms, and the GW discrepancy provides a pseudometric to measure their reconstruction errors. 
We design an effective approximate algorithm for learning the atoms and the graph embeddings (the weights of the atoms), unrolling loopy computations of GW discrepancy and barycenters and simplifying backpropagation based on the envelope theorem~\cite{afriat1971theory}. 
The approximate algorithm can be implemented based on either proximal point algorithm (PPA)~\cite{xu2019scalable} or Bregman alternating direction method of multipliers (BADMM)~\cite{wang2014bregman}. 

Our GWF model explicitly factorizes graphs into a set of atoms. 
The atoms are shared by all observations while their weights specialized for different individuals. 
This model has several advantages. 
Firstly, it is with high flexibility. 
The observed graphs, the atoms, and the GW barycenters can be with different sizes, and their alignment is achieved by the optimal transport corresponding to the GW discrepancy between them. 
Secondly, this model is compatible with existing models, which can be learned based on backpropagation and can be used as a structural regularizer in supervised learning. 
In the aspect of data, this model is applicable no matter whether the graphs are with node attributes or not.
Thirdly, the graph embeddings derived by our model is more interpretable --- they directly reflect the significance of the atoms.
To our knowledge, our work makes the first attempt to establish an explicit factorization mechanism for graphs, which extends traditional factorization models under GW discrepancy. 
Experimental results show that the GWF model achieves encouraging results in the task of graph clustering.

\section{Proposed Model}
In this work, we represent a graph as its adjacency matrix $\bm{C}\in\mathbb{R}^{N\times N}$, whose elements are nonnegative. 
Optionally, when the graph is with $D$-dimensional node attributes, we represent its node attributes as a matrix $\bm{F}\in\mathbb{R}^{N\times D}$. 
Furthermore, for a graph we denote the empirical distribution of its nodes as $\bm{\mu}\in \Delta^{N-1}$, where $\Delta^{N-1}=\{\bm{x}=[x_n]\in\mathbb{R}^N | x_n\geq 0,~\text{and}~\sum_k x_k=1\}$ represents a $(N-1)$-simplex. 
Following~\cite{peyre2016gromov}, we assume that the empirical distribution is uniform, $i.e.$, $\bm{\mu}=\frac{1}{N}\bm{1}_N$.\footnote{Our model is applicable for other distributions, $e.g.$, the distribution derived from node degree~\cite{xu2019scalable}.} 
Given a set of observations $\{\bm{C}_i\in [0, \infty)^{N_i\times N_i},\bm{\mu}_i\in \Delta^{N_i-1}\}_{i=1}^{I}$, we aim at designing a factorization model with $K$ atoms $\bm{U}_{1:K}=\{\bm{U}_k=[u_{ij}^k]\in [0, \infty)^{N_k\times N_k}\}_{k=1}^{K}$ and representing each observation $\bm{C}_i$ as an embedding vector $\bm{\lambda}_i=[\lambda_{ik}]\in \mathbb{R}^{K}$, such that the element $\lambda_{ik}$ can be interpreted as the significance of the $k$-th atom for the $i$-th observation. 
Here, we assume that each $\bm{\lambda}_i$ is in a $(K-1)$-simplex as well, $i.e.$, $\bm{\lambda}_i\in\Delta^{K-1}$ for $i=1,...,I$.
It should be noted that generally for different $\bm{C}_i$ and $\bm{C}_j$, their nodes are unaligned and $N_i\neq N_j$.

\subsection{Revisiting factorization models}
Many existing factorization models assume that each vectorized data $\bm{y}\in\mathbb{R}^N$ can be represented as the weighted sum of $K$ atoms, $i.e.$, $\bm{y}=\sum_{k}\lambda_k\bm{a}_k=\bm{A\lambda}$, where $\bm{A}=[\bm{a}_k]\in\mathbb{R}^{N\times K}$ contains $K$ atoms and $\bm{\lambda}=[\lambda_{k}]\in\mathbb{R}^K$ is the latent representation of $\bm{y}$. 
Given a set of observations $\{\bm{y}_i\}_{i=1}^{I}$, a straightforward method to learn a factorization model is solving the following optimization problem: 
\begin{eqnarray}\label{eq:fm}
\begin{aligned}
\sideset{}{_{\{\bm{A},\bm{\lambda}_{1:I}\}\in\Omega}}\min\sideset{}{_{i=1}^{I}}\sum d_{\text{loss}}^p(\bm{A}\bm{\lambda}_i, \bm{y}_i).
\end{aligned}
\end{eqnarray}
Here, $\Omega$ represents the constraints on $\bm{A}$ and/or $\bm{\lambda}_{1:I}$.  $d_{\text{loss}}(\cdot,\cdot)$ defines the distance between $\bm{y}_i$ and its estimation $\bm{A}\bm{\lambda}_i$, which is used as the loss function, and $p$ indicates the order of $d_{\text{loss}}$. 
This optimization problem can be specialized as many classical models: without any constraints, PCA~\cite{pearson1901liii} sets $d_{\text{loss}}$ to Euclidean distance and $p=2$; robust PCA~\cite{candes2011robust} sets $d_{\text{loss}}$ to $l_1$-norm and $p=1$; NMF~\cite{sra2006generalized} sets $\Omega$ to the set of nonnegative matrices; 
the Wasserstein dictionary learning~\cite{rolet2016fast} sets $d_{\text{loss}}$ to Wasserstein distance and $\Omega$ the set of the nonnegative matrices with normalized columns.

Furthermore, when $\bm{\lambda}\in\Delta^{K-1}$, the linear factorization model $\bm{A\lambda}$ corresponds to learning a barycenter of the atoms in the Euclidean space, which can be rewritten as
\begin{eqnarray}
\begin{aligned}
\bm{b}(\bm{A},\bm{\lambda}; d_{2}^2):=\bm{A\lambda}=\arg\sideset{}{_{\bm{y}}}\min\sideset{}{_{k=1}^{K}}\sum \lambda_k\| \bm{y} - \bm{a}_k \|_2^2.
\end{aligned}
\end{eqnarray}
Here, $\bm{y}$ represents the variable of the optimization problem and its optimal solution is $\bm{A\lambda}$ ($i.e.$, an estimation of $\bm{y}_i$ in (\ref{eq:fm})).
Extending the Euclidean metric to other metrics, we obtain nonlinear factorization models, $i.e.$, $\bm{b}(\bm{A},\bm{\lambda};d_{\text{b}}^q):=\min_{\bm{y}}\sum_{k=1}^{K}\lambda_k d_{\text{b}}^{q}(\bm{y}, \bm{a}_k)$, where $d_{\text{b}}(\cdot,\cdot)$ is the metric used to calculate barycenters and $q$ is its order.
For example, when $\bm{a}_k\in \Delta^{N-1}$ for $k=1,...,K$ and $d_{\text{b}}$ is Wasserstein metric, $\bm{b}(\bm{a}_{1:K},\bm{\lambda};d_{\text{b}}^q)$ corresponds to the Wasserstein factorization models~\cite{schmitz2018wasserstein}. 
In summary, many existing factorization models are in the following framework:
\begin{eqnarray}\label{eq:gmf_loss}
\sideset{}{_{\{\bm{A},\bm{\lambda}_{1:I}\}\in\Omega}}\min\sideset{}{_{i=1}^{I}}\sum d_{\text{loss}}^p(\bm{b}(\bm{A},\bm{\lambda}_i;d_{\text{b}}^q), \bm{y}_i).
\end{eqnarray}

\subsection{Gromov-Wasserstein factorization}
As aforementioned, when the observed data are graphs, $i.e.$, replacing the vectors $\bm{y}_{1:I}$ and the atoms $\bm{A}$ in (\ref{eq:gmf_loss}) with graphs $\bm{C}_{1:I}$ and their atoms $\bm{U}_{1:K}$, respectively, the classic metrics become inapplicable. 
In such a situation, we set the $d_{\text{loss}}$ and $d_{\text{b}}$ in (\ref{eq:gmf_loss}) to GW discrepancy, denoted as $d_{gw}$, and achieve the proposed GWF model. 
The GW discrepancy is an extension of the Gromov-Wasserstein distance~\cite{memoli2011gromov} on metric measure spaces:

\noindent \textbf{Definition} \emph{(Gromov-Wasserstein distance)
Let $(X, d_X, u_X)$ and $(Y, d_Y, u_Y)$ be two metric measure spaces, where $(X, d_X)$ is a compact metric space and $u_X$ is a Borel probability measure on $X$ (with $(Y, d_Y, u_Y)$ defined in the same way). 
For $p\in[1,\infty)$, the $p$-th order Gromov-Wasserstein distance $d_{gw}(u_X, u_Y)$ is defined as 
\begin{eqnarray*}
\inf_{\pi\in \Pi(u_X, u_Y)}\Bigl(\iint_{X\times Y, X\times Y} L_{x, y, x', y'}^p\,d\pi(x, y)\,d\pi(x',y')\Bigr)^{\frac{1}{p}},
\end{eqnarray*}
where $L_{x, y, x', y'}=|d_X(x, x')-d_Y(y,y')|$ is the loss function and $\Pi(u_X,u_Y)$ is the set of all probability measures on $X\times Y$ with $u_X$ and $u_Y$ as marginals.}

The GW discrepancy is similar to the GW distance, but it does not require the $d_{X}$ and the $d_Y$ to be strict metrics. 
Therefore, it defines a flexible pseudometric on structural data like graphs~\cite{chowdhury2018gromov,xu2019scalable} and point clouds~\cite{peyre2016gromov}. 
In particular, given two graphs $\{\bm{C}_s=[c_{ij}^s]\in\mathbb{R}^{N_s\times N_s}, \bm{C}_t=[c_{i'j'}^t]\in\mathbb{R}^{N_t\times N_t}\}$ and their empirical entity distributions $\{\bm{\mu}_s\in\mathbb{R}^{N_s},\bm{\mu}_t\in\mathbb{R}^{N_t}\}$, the order-2 GW discrepancy between them, denoted as $d_{gw}(\bm{C}_s, \bm{C}_t)$, is defined as
\begin{eqnarray}\label{eq:gwd}
\min_{\bm{T}\in \Pi(\bm{\mu}_s,\bm{\mu}_t)}\Bigl(
\sideset{}{_{i,j=1}^{N_s}}\sum\sideset{}{_{i',j'=1}^{N_t}}\sum |c_{ij}^s - c_{i'j'}^t|^{2}T_{ii'}T_{jj'}
\Bigr)^{\frac{1}{2}},
\end{eqnarray}
where $\Pi(\bm{\mu}_s,\bm{\mu}_t)=\{\bm{T}\geq\bm{0}|\bm{T}\bm{1}_{N_t}=\bm{\mu}_s,\bm{T}^{\top}\bm{1}_{N_s}=\bm{\mu}_t\}$. 
The optimal $\bm{T}$ indicates the optimal transport between the nodes of $\bm{C}_s$ and those of $\bm{C}_t$.
According to~\cite{peyre2016gromov}, we can rewrite (\ref{eq:gwd}) as 
\begin{eqnarray}\label{eq:gwd2}
d_{gw}(\bm{C}_s,\bm{C}_t):=\min_{\bm{T}\in \Pi(\bm{\mu}_s,\bm{\mu}_t)} (\langle \bm{C}_{st} - 2\bm{C}_s\bm{T}\bm{C}_t^{\top},~\bm{T} \rangle)^{\frac{1}{2}},
\end{eqnarray}
where $\langle \cdot,\cdot\rangle$ represents the inner product of matrices, $\bm{C}_{st}
=(\bm{C}_s\odot \bm{C}_s)\bm{\mu}_s\bm{1}_{N_t}^{\top}+\bm{1}_{N_s}\bm{\mu}_t^{\top}(\bm{C}_t\odot\bm{C}_t)^{\top}$ and $\odot$ represents the Hadamard product of matrices.
The computation of the GW discrepancy corresponds to an optimal transport problem. 
This problem can be solved iteratively by the entropic regularization-based method~\cite{peyre2016gromov} or the proximal point algorithm~\cite{xu2019scalable}. 

Given $K$ graphs $\{\bm{U}_{k}\}_{k=1}^{K}$ and their weights $\bm{\lambda}=[\lambda_k]\in\Delta^{K-1}$, we can naturally define their order-2 GW barycenter based on the GW discrepancy in (\ref{eq:gwd2}):
\begin{eqnarray}\label{eq:gwb}
\bm{B}(\bm{U}_{1:K},\bm{\lambda};d_{gw}^2):=\arg\sideset{}{}\min_{\bm{B}}\sideset{}{_{k}}\sum\lambda_k d_{gw}^2(\bm{B}, \bm{U}_{k}).
\end{eqnarray}
According to the definition we can find that the computation of the GW barycenter involves solving $K$ optimal transport problems iteratively~\cite{peyre2016gromov}.

Plugging $d_{gw}$ into (\ref{eq:gmf_loss}) and setting $p=q=2$, we obtain the proposed  Gromov-Wasserstein factorization model:
\begin{eqnarray}\label{eq:gwf}
\min_{\bm{U}_{1:K}\geq\bm{0},\bm{\lambda}_{1:I}\in\Delta^{K-1}} \sideset{}{_{i=1}^{I}}\sum d_{gw}^2(\bm{B}(\bm{U}_{1:K},\bm{\lambda}_i;d_{gw}^2), \bm{C}_i).
\end{eqnarray}
In this model, each graph $\bm{C}_i$ is estimated by a GW barycenter of atoms $\bm{U}_{1:K}$.
The weights associated with the atoms formulate the embedding vector $\bm{\lambda}_i$.
Both the atoms and the embeddings are learned to minimize the GW discrepancy between the observed graphs and their GW barycenter-based estimations.
As aforementioned, we require the elements of each atom to be nonnegative and make each embedding in a $(K-1)$-simplex.

As shown in Figure~\ref{fig:scheme}, this GWF model applies several atoms to construct a collection of graphs, in which each graph is a barycenter of the atoms. 
Each observed graph is reconstructed by the most similar graph in the collection, and the embedding vector $\bm{\lambda}_i$ corresponding to the reconstructed graph indicates the significance of different atoms. 
The embeddings of observed graphs can be used as features for many downstream tasks, $e.g.$, graph clustering.
The GWF model is very flexible --- the atoms and the observed graphs can be with different sizes.
Differing from the graph embedding methods in~\cite{henaff2015deep,ying2018hierarchical}, the proposed GWF model does not rely on any side information like labels of graphs in the learning phase. 
Additionally, because of using an explicit factorization mechanism, the proposed model has better interpretability than many existing methods.

\section{Learning Algorithm}
\subsection{Reformulation of the problem}
Learning GWF models is challenging because (\ref{eq:gwf}) is a highly-nonlinear constrained optimization problem. 
As shown in~(\ref{eq:gwd2},~\ref{eq:gwb}), the computation of the GW discrepancy and that of the GW barycenter correspond to two optimization problems, respectively, so the GWF model in (\ref{eq:gwf}) is a complicated composition of multiple optimization tasks that are with different variables. 
Facing such a difficult problem, we reformulate it and simplify its computations.

\textbf{Reparametrization of target variables} 
To reformulate (\ref{eq:gwf}) as an unconstrained problem, we further parametrize the parameters $\bm{U}_{1:K}$ and $\bm{\lambda}_{1:I}$ as
\begin{eqnarray}\label{eq:func}
\begin{aligned}
&u_{ij}^k = f(v_{ij}^k),~\forall k=1,...,K,~\forall i,j=1,...,N_k,\\
&\bm{\lambda}_i = g(\bm{z}_i),~\forall i=1,...,I.
\end{aligned}
\end{eqnarray}
where $\bm{V}_{1:K}=\{\bm{V}_k=[v_{ij}^k]\in\mathbb{R}^{N_k\times N_k}\}_{k=1}^{K}$ and $\bm{z}_{1:I}=\{\bm{z}_i=[z_{ik}]\in\mathbb{R}^{K}\}$ are new unconstrained parameters, $f(\cdot)=\text{ReLU}(\cdot)$ and $g(\cdot)=\text{Softmax}(\cdot)$ are two functions mapping the new parameters to the feasible domains of original problem. 
Note that we use $\text{ReLU}(\cdot)$ to pursue atoms with sparse edges.
Accordingly, we use the $\bm{z}_i$ as the embedding of the graph $\bm{C}_i$.

\begin{figure}[!t]
    \centering
    \subfigure[Stacked computational modules]{
    \includegraphics[height=1.8cm]{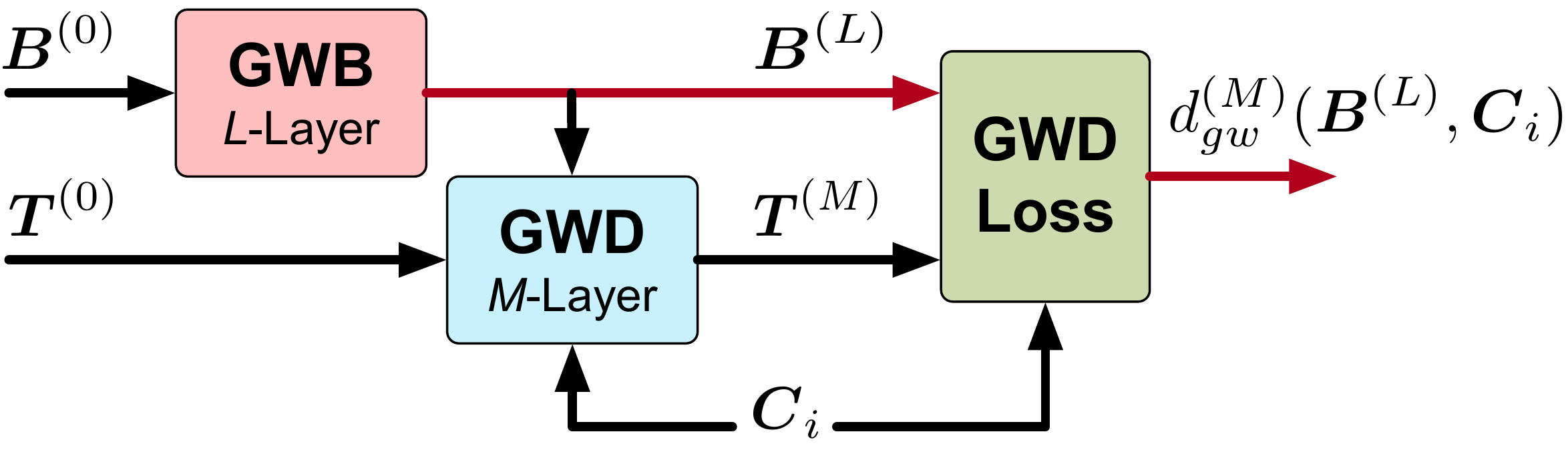}\label{fig:alg_gwf}
    }\\
    \subfigure[One layer of the GW barycenter module]{
    \includegraphics[height=2.8cm]{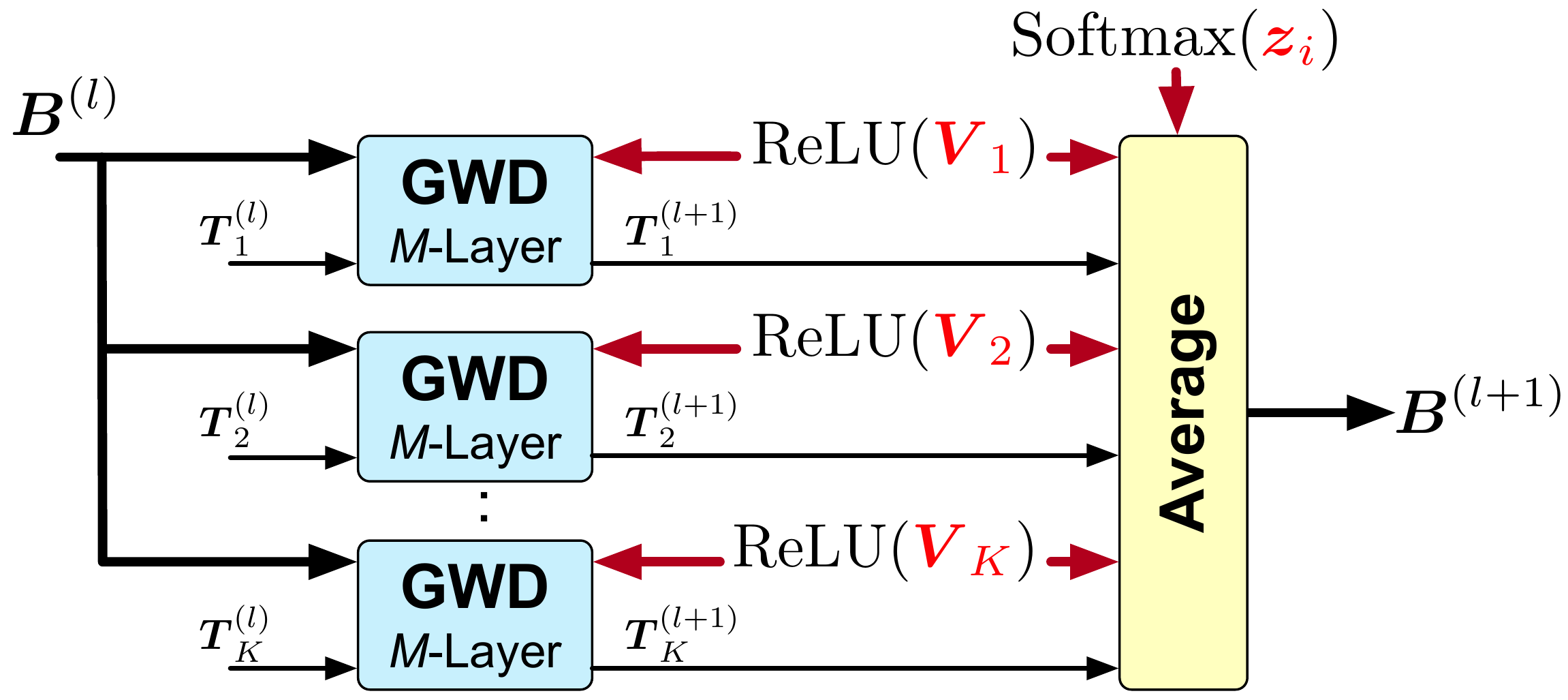}\label{fig:alg_gwb}
    }\\
    \caption{
    \small{(a) The stacked modules for learning GWF models. 
    (b) One layer of the GWB module. 
    In each figure, the red arrows represent the paths used for both feedforward computation and backpropagation, while the black ones are just for feedforward computation. 
    The GWD corresponds to Algorithms~\ref{alg:ot-ppa} and~\ref{alg:ot-badmm},  the GWB corresponds to Algorithm~\ref{alg:gwb}, and the GWD loss corresponds to (\ref{eq:approx}).}
    }
    \label{fig:algs}
\end{figure}

\textbf{Unrolling loops} 
For each $\bm{C}_i$, the feedforward computation of its objective function in (\ref{eq:gwf}) corresponds to two steps: $i$) solving $K$ optimal transport problems iteratively to estimate a GW barycenter; $ii$) solving one more optimal transport problem to derive the GW discrepancy between the GW barycenter and $\bm{C}_i$. 
Each step contains loopy computations of optimal transport matrices. 
We unroll the loops as stacked modules shown in Figure~\ref{fig:alg_gwf}. 
Specifically, we design a Gromov-Wasserstein discrepancy (GWD) module with $M$ layers to achieve an approximation of GW discrepancy. 
Based on this GWD module, we further propose a GW barycenter (GWB) module with $L$ layers to obtain an approximation of GW barycenter. 
Figure~\ref{fig:alg_gwb} illustrates one layer of the GWB module.

Based on these two modifications, we reformulate (\ref{eq:gwf}) as
\begin{eqnarray}
\min_{\bm{V}_{1:K},\bm{z}_{1:I}} \sideset{}{_{i=1}^{I}}\sum d_{gw}^{2(M)}(\bm{B}^{(L)}(f(\bm{V}_{1:K}),g(\bm{z}_i);d_{gw}^2), \bm{C}_i).
\end{eqnarray}

\begin{algorithm}[t]
\small{
	\caption{$\text{GWD}_{\text{PPA}}^{(M)}(\bm{C}_s,\bm{C}_t)$}
	\label{alg:ot-ppa}
	\begin{algorithmic}[1]
	    \STATE $\bm{C}_{st}=(\bm{C}_s\odot\bm{C}_s)\bm{1}\bm{\mu}_t^{\top}+\bm{\mu}_s\bm{1}^{\top}(\bm{C}_t\odot\bm{C}_t)^{\top}$
	    \STATE $\bm{T}^{(0)}=\bm{\mu}_s\bm{\mu}_t^{\top}$, $\bm{a}=\bm{\mu}_s$
		\STATE \textbf{for} $m=0,...,M-1$ 
		\STATE \quad $\bm{\Phi}=\exp(-\frac{1}{\gamma}(\bm{C}_{st}-2\bm{C}_s\bm{T}^{(m)}\bm{C}_t^{\top}))\odot\bm{T}^{(m)}$.
		\STATE \quad $\bm{b}=\frac{\bm{\mu}_t}{\bm{\Phi}^{\top}\bm{a}}$,  $\bm{a} = \frac{\bm{\mu}_s}{\bm{\Phi}\bm{b}}$, and 
		$\bm{T}^{(m\text{+}1)} = \text{diag}(\bm{a})\bm{\Phi}\text{diag}(\bm{b})$.
		\RETURN $\bm{T}^{(M)}$
	\end{algorithmic}
}
\end{algorithm}

\subsection{Implementations of the modules}
The GWD module is the backbone of our algorithm. 
In this paper, we propose two options to implement this module. 
This first one is the proximal point algorithm (PPA) in~\cite{xu2019gromov,xu2019scalable}.
Given $\{\bm{C}_s,\bm{\mu}_s\}$ and $\{\bm{C}_t,\bm{\mu}_t\}$, the PPA solve (\ref{eq:gwd2}) iteratively. 
In each iteration, it solves the following problem:
\begin{eqnarray*}\label{eq:dgw1}
\begin{aligned}
\bm{T}^{(m\text{+}1)}=&\arg\sideset{}{_{\bm{T}\in \Pi(\bm{\mu}_s,\bm{\mu}_t)}}\min\langle \bm{C}_{st} - 2\bm{C}_s\bm{T}^{(m)}\bm{C}_t^{\top},\bm{T}\rangle\\
&+ \gamma\text{KL}(\bm{T}\lVert\bm{T}^{(m)}),
\end{aligned}
\end{eqnarray*}
where $\text{KL}(\bm{T}\lVert\bm{T}^{(m)})$ computes the KL-divergence between the optimal transport matrix and its previous estimation. 
We solve this problem approximately by one-step Sinkhorn-Knopp update~\cite{sinkhorn1967concerning,xu2019scalable}. 
After $M$ iterations, we derive the $M$-step approximation of the optimal transport matrix. 
Accordingly, we show the scheme of the PPA-based GWD module in Algorithm~\ref{alg:ot-ppa}.
This method is based on the work in~\cite{xu2019gromov}, which replaces the entropy regularizer in~\cite{peyre2016gromov} with a KL divergence. 
According to~\cite{xu2019gromov,xu2019scalable}, the PPA outperforms the entropic GW method~\cite{peyre2016gromov} on both stability and convergence.

Besides the PPA-based GWD module, we propose a different kind of GWD module based on the Bregman alternating direction method of multipliers (BADMM)~\cite{wang2014bregman}. 
Specifically, introducing an auxiliary variable $\bm{S}$, we rewrite (\ref{eq:gwd2}) as
\begin{eqnarray}\label{eq:dgw2}
\sideset{}{_{\bm{T}\in \Pi(\bm{\mu}_s,\cdot),\bm{S}\in \Pi(\cdot,\bm{\mu}_t),\bm{T}=\bm{S}}}\min\langle \bm{C}_{st} - 2\bm{C}_s\bm{S}\bm{C}_t^{\top},~\bm{T}\rangle,
\end{eqnarray}
where $\Pi(\bm{\mu}_s,\cdot)=\{\bm{T}\geq \bm{0}~|~\bm{T}\bm{1}_M=\bm{\mu}_s\}$ and $\Pi(\cdot, \bm{\mu}_t)=\{\bm{T}\geq \bm{0}~|~\bm{T}^{\top}\bm{1}_N=\bm{\mu}_t\}$. 
We further introduce a dual variable $\bm{Z}$ and solve (\ref{eq:dgw2}) by the following three steps~\cite{wang2014bregman,ye2017fast}.
\begin{eqnarray}\label{eq:updateP2}
\begin{aligned}
\bm{T}^{(m\text{+}1)}=&\arg\sideset{}{_{\bm{T}\in \Pi(\bm{\mu}_s,\cdot)}}\min\langle \bm{C}_{st} - 2\bm{C}_s\bm{S}^{(m)}\bm{C}_t^{\top},\bm{T}\rangle\\
&+\langle\bm{Z}^{(m)},\bm{T}-\bm{S}^{(m)}\rangle+\gamma\text{KL}(\bm{T}\lVert\bm{S}^{(m)}),\\
\bm{S}^{(m\text{+}1)}=&\arg\sideset{}{_{\bm{S}\in \Pi(\cdot,\bm{\mu}_t)}}\min\langle - 2\bm{C}_s^{\top}\bm{T}^{(m\text{+}1)}\bm{C}_t,\bm{S}\rangle\\
&+\langle\bm{Z}^{(m)},\bm{T}^{(m\text{+}1)}-\bm{S}\rangle+\gamma\text{KL}(\bm{S}\lVert\bm{T}^{(m\text{+}1)}),\\
\bm{Z}^{(m\text{+}1)} = &\bm{Z}^{(m)} + \gamma(\bm{T}^{(m\text{+}1)}-\bm{S}^{(m\text{+}1})).
\end{aligned}
\end{eqnarray}
Accordingly, Algorithm~\ref{alg:ot-badmm} gives the scheme of the BADMM-based GWD module.

\begin{algorithm}[t]
\small{
	\caption{$\text{GWD}_{\text{BADMM}}^{(M)}(\bm{C}_s,\bm{C}_t)$}
	\label{alg:ot-badmm}
	\begin{algorithmic}[1]
	    \STATE $\bm{C}_{st}=(\bm{C}_s\odot\bm{C}_s)\bm{1}\bm{\mu}_t^{\top}+\bm{\mu}_s\bm{1}^{\top}(\bm{C}_t\odot\bm{C}_t)^{\top}$
	    \STATE $\bm{T}^{(0)}=\bm{\mu}_s\bm{\mu}_t^{\top}$, $\bm{Z}=\bm{0}$.
		\STATE \textbf{for} $m=0,...,M-1$ 
		\STATE \quad $\bm{\Phi}_1=\exp(\frac{1}{\gamma}(2\bm{C}_s^{\top}\bm{T}^{(m)}\bm{C}_t+\bm{Z}))\odot\bm{T}^{(m)}$.
		\STATE \quad $\bm{S}=\bm{\Phi}_1\text{diag}(\frac{\bm{\mu}_t}{\bm{\Phi}_1^{\top}\bm{1}})$.
		\STATE \quad $\bm{\Phi}_2=\exp(-\frac{1}{\gamma}(\bm{C}_{st}-2\bm{C}_s\bm{S}\bm{C}_t^{\top}+\bm{Z}))\odot \bm{S}$.
		\STATE \quad $\bm{T}^{(m\text{+}1)} = \text{diag}(\frac{\bm{\mu}_s}{\bm{\Phi}_2\bm{1}})\bm{\Phi}_2$.
		\STATE \quad $\bm{Z} = \bm{Z} +\gamma(\bm{T}^{(m\text{+}1)}-\bm{S})$.
		\RETURN $\bm{T}^{(M)}$
	\end{algorithmic}
}
\end{algorithm}

\begin{figure}[!t]
    \centering
    \subfigure[Undirected graphs]{
    \includegraphics[height=3.8cm]{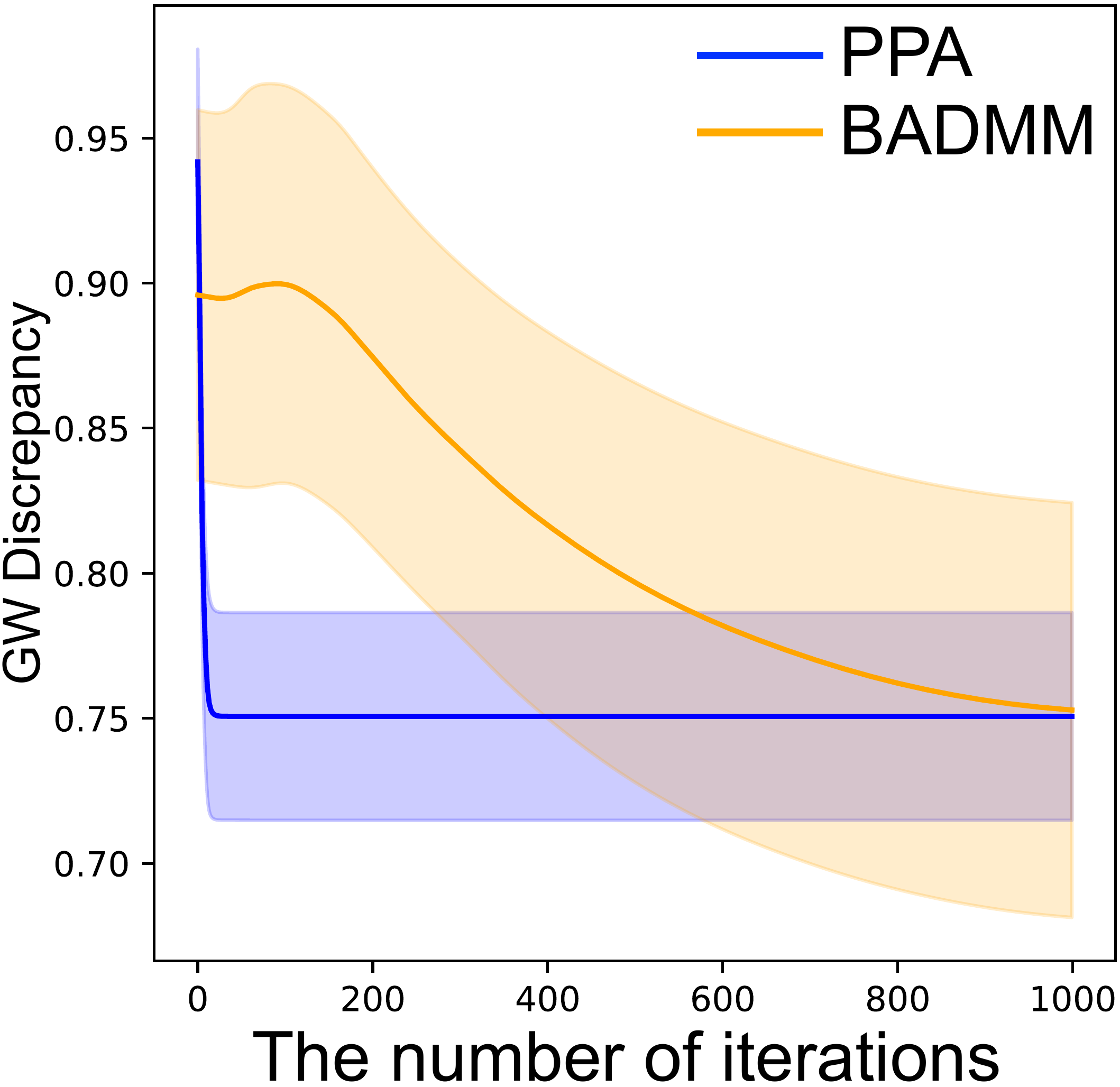}\label{fig:undirected}
    }
    \subfigure[Directed graphs]{
    \includegraphics[height=3.8cm]{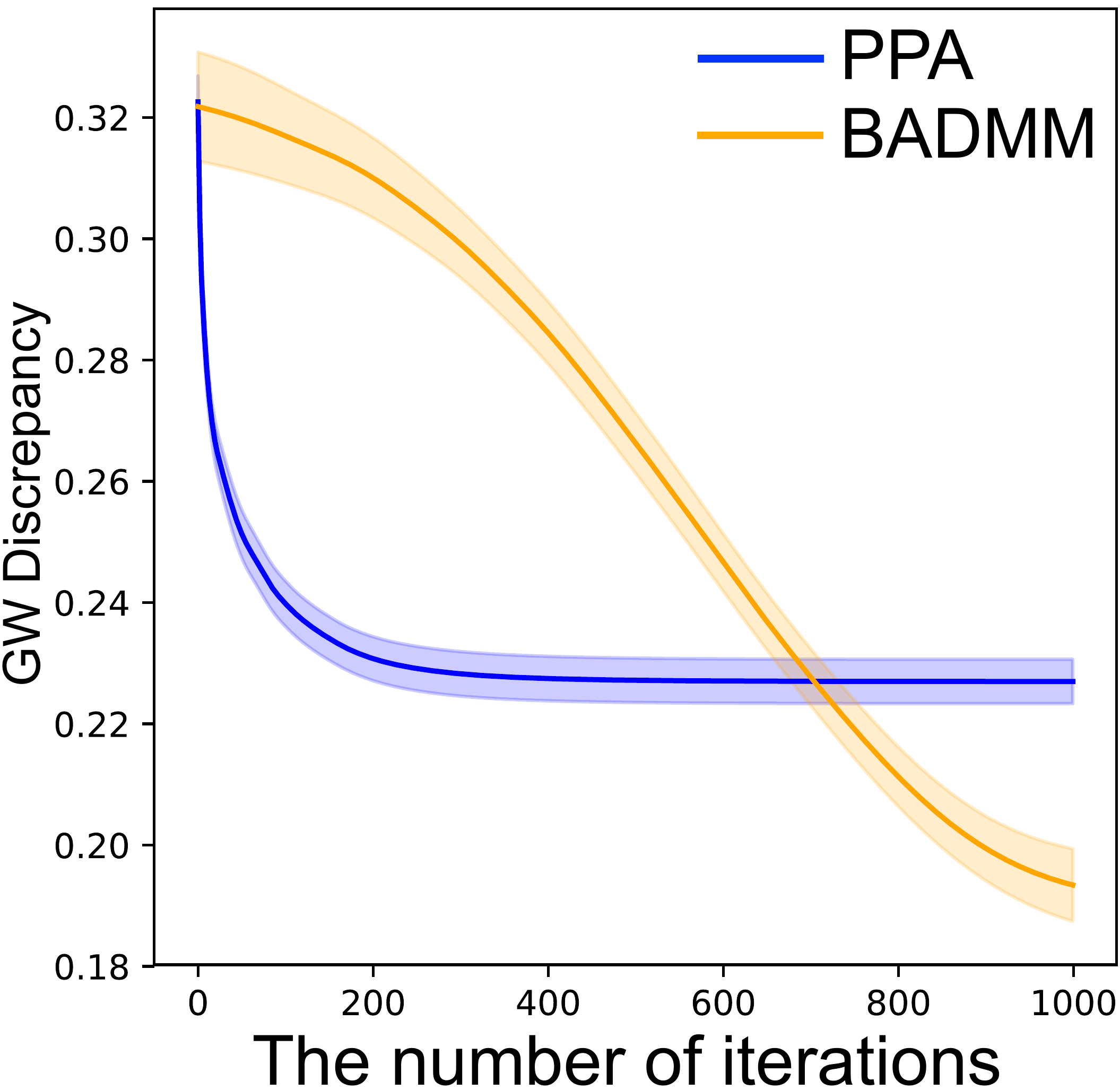}\label{fig:directed}
    }
    \caption{
    \small{Comparisons for PPA and BADMM.}
    }
    \label{fig:cmp}
\end{figure}

The BADMM algorithm is originally designed for computing Wasserstein distance~\cite{wang2014bregman,ye2017fast}. 
To our knowledge, our work is the first attempt to apply BADMM to compute the GW discrepancy. 
We test these two GWD modules on synthetic graphs and find that they are suitable for different scenarios. 
In particular, we synthesize 100 pairs of undirected graphs and 100 pairs of directed graphs, respectively.
Each graph is with 100 nodes.
The directed graphs are generated based on Barab{\'a}si-Albert (BA) model~\cite{barabasi2016network}. 
For each directed graph, we add its adjacency matrix to its transpose and derive an undirected graph accordingly. 
For each pair of graphs, we apply our GWD modules to compute the GW discrepancy between the two graphs, and then, we calculate the mean and the standard deviation of the GW discrepancy in each step. 
The comparisons for the PPA-based module and the BADMM-based module are shown in Figures~\ref{fig:undirected} and \ref{fig:directed}.
The PPA-based module requires fewer steps than the BADMM-based module to converge to a stable optimal transport matrix for both undirected and directed graphs. 
However, the BADMM-based module can achieve smaller GW discrepancy when applying to directed graphs. 
In other words, we need to select different GWD modules according to the structures of observed graphs and the practical constraints on computational complexity.

The GWB module is implemented based on the GWD modules. 
As shown in Algorithm~\ref{alg:gwb}, given an initial GW barycenter $\bm{B}^{(0)}$ and its empirical distribution $\bm{\mu}_b$, we obtain a $L$-step approximation of GW barycenter and the optimal transport matrices between the barycenter and the atoms. 
In particular, we calculate the GW barycenter via alternating optimization: 1) update the optimal transports between the atoms and the barycenter; and 2) update the barycenter accordingly. 
Figure~\ref{fig:alg_gwb} illustrates one step of the GWB module ($i.e.$, the lines 2-4 in Algorithm~\ref{alg:gwb}), where $\bm{T}_k^{(l)}$ is the estimated optimal transport in the $l$-th step of the GWB module. 
We update it by calling a GWD-module with $M$ inner iterations, and the output is denoted as $\bm{T}_k^{(l+1)}$.

\begin{algorithm}[t]
\small{
	\caption{$\text{GWB}^{(L)}(\bm{U}_{1:K},\bm{\lambda},\bm{B}^{(0)},\bm{\mu}_b)$}
	\label{alg:gwb}
	\begin{algorithmic}[1]
	    \STATE \textbf{for} $l=0,...,L-1$
	    \STATE \quad \textbf{for} $k=1,...,K$
	    \STATE \quad\quad $\bm{T}_k^{(l\text{+}1)}= \text{GWD}^{(M)}(\bm{B}^{(l)},\bm{U}_k)$.
	    \STATE \quad $\bm{B}^{(l\text{+}1)} =\frac{1}{\bm{\mu}_b\bm{\mu}_b^{\top}} \sum_{k=1}^{K}\lambda_k\bm{T}_k^{(l\text{+}1)}\bm{U}_k(\bm{T}_k^{(l\text{+}1)})^{\top}$.
	    \RETURN $\bm{B}^{(L)}$, $\bm{T}_{1:K}^{(L)}$
	\end{algorithmic}
}
\end{algorithm}

\subsection{Envelope theorem-based backpropagation}
Given the modules proposed above, we can compute $d_{gw}^{2(M)}(\bm{B}^{(L)}(f(\bm{V}_{1:K}),g(\bm{z}_i);d_{gw}^2), \bm{C}_i)$ easily and apply backpropagation to update variables $\bm{V}_{1:K}$ and $\bm{z}_{1:L}$. 
Here, we take advantage of the envelope theorem~\cite{afriat1971theory} to simplify the computation of backpropagation. 
In particular, we compute each optimal transport matrix, $i.e.$, the $\bm{T}^{(L)}_{1:K}$ used to calculate the GW barycenter and the $\bm{T}^{(M)}$ between the GW barycenter and the observed graph, based on current parameters. 
Accordingly, the objective function becomes
\begin{eqnarray}\label{eq:approx}
\begin{aligned}
&d_{gw}^{2(M)}(\bm{B}^{(L)}(f(\bm{V}_{1:K}),g(\bm{z}_i);d_{gw}^2), \bm{C}_i)\\
&=\langle \bm{C}_{st}-2\bm{B}^{(L)}\bm{T}^{(M)}\bm{C}_i^{\top},~\bm{T}^{(M)} \rangle
\end{aligned}
\end{eqnarray}
where
\begin{eqnarray}
\begin{aligned}
&\bm{C}_{st} = (\bm{B}^{(L)}\odot \bm{B}^{(L)})\bm{\mu}_b \bm{1}_{N_i}^{\top}+\bm{1}_{N_i}\bm{\mu}_i^{\top}(\bm{C}_i\odot\bm{C}_i)^{\top},\\
&\bm{B}^{(L)} = \frac{1}{\bm{\mu}_b\bm{\mu}_b^{\top}} \sideset{}{_{k=1}^{K}}\sum g_k(\bm{z}_i)\bm{T}_k^{(L)}f(\bm{V}_k)(\bm{T}_k^{(L)})^{\top}.
\end{aligned}
\end{eqnarray}
Based on the envelope theorem above, we calculate the gradient of (\ref{eq:approx}) with fixed optimal transport matrices. 
The gradients for the optimal transport matrices can be ignored when applying backpropagation, which improves the efficiency of our algorithm significantly. 
In summary, we learn our GWF model by Algorithm~\ref{alg:gwf}.
\begin{algorithm}[t]
\small{
 	\caption{Learning GWF models}
 	\label{alg:gwf}
 	\begin{algorithmic}[1]
 		\REQUIRE A dataset $\mathcal{D}=\{\bm{C}_i,\bm{\mu}_i\}_{i=1}^{I}$.
 		\STATE Initialize atoms $\bm{V}_{1:K}$ and $\bm{z}_{1:I}$ randomly.
 		\STATE Set the node distributions of atoms as uniform distributions.
 		\STATE For {$\text{epoch}=1, 2, ...$}
 		\STATE \quad For {$i = 1, ..., I$}
 	    \STATE \quad\quad Initialize $\bm{B}^{(0)} = \bm{C}_i$, $\bm{\mu}_b = \bm{\mu}_i$.
 	    \STATE \quad\quad $\bm{B}^{(L)},\bm{T}_{1:K}^{(L)}$=$ \text{GWB}^{(L)}(f(\bm{V}_{1:K}),g(\bm{z}_i),\bm{B}^{(0)},\bm{\mu}_b)$.
 	    \STATE \quad\quad $\bm{T}^{(M)}= \text{GWD}^{(M)}(\bm{B}^{(L)},\bm{C}_i)$.
 	    \STATE \quad\quad Get $d_{gw}^{2(M)}(\bm{B}^{(L)}(f(\bm{V}_{1:K}),g(\bm{z}_i);d_{gw}^2), \bm{C}_i)$ via (\ref{eq:approx}).
 	    \STATE \quad\quad Update $\bm{V}_{1:K}$ and $\bm{z}_i$ via backpropagation.
 	    \RETURN $\bm{V}_{1:K}$ and $\bm{z}_{1:I}$.
 	\end{algorithmic}
}
\end{algorithm}

\subsection{Extensions}
Given more side information, we can extend our GWF models to more complicated scenarios. 

\textbf{Learning with labels} When some graphs are labeled, we can achieve semi-supervised learning of our GWF model by adding a label-related loss:
\begin{eqnarray}\label{eq:ssl}
\begin{aligned}
\min_{\bm{V}_{1:K},\bm{z}_{1:I},\phi}& \sideset{}{_{i=1}^{I}}\sum d_{gw}^{2(M)}(\bm{B}^{(L)}(f(\bm{V}_{1:K}),g(\bm{z}_i); d_{gw}^2), \bm{C}_i)\\
&+ \beta \sideset{}{_{i\in\mathcal{D}_l}}\sum\text{loss}(\phi(\bm{z}_i), l_i).
\end{aligned}
\end{eqnarray}
Here, $\text{loss}(\phi(\bm{z}_i), l_i)$ represents the label-related loss, where $l_i$ is the label of the $i$-th graph and $\phi(\cdot)$ is a learnable function mapping the embedding $\bm{z}_i$ to the label space. 
A typical choice of the loss is $\text{CrossEntropy}(\text{MLP}(\bm{z}_i)), l_i)$. 

\textbf{Learning with node attributes}
Sometimes the nodes of each graph are associated with vectorized features. 
In such a situation, the observations are $\{\bm{C}_i, \bm{\mu}_i, \bm{F}_i\}_{i=1}^{I}$, where $\bm{F}_i\in\mathbb{R}^{N_i\times D}$ represents the features of nodes. 
Accordingly, the atoms of our GWF model becomes $\{\bm{U}_{1:K}, \bm{H}_{1:K}\}$, where $\bm{U}_{k}\in [0,\infty)^{N_k\times N_k}$ is the adjacency matrix of the $k$-th atom and $\bm{H}_k\in\mathbb{R}^{N_k\times D}$ represents the features of its nodes. 
To learn this GWF model, we can replace the GW discrepancy with the fused Gromov-Wasserstein (FGW) discrepancy in~\cite{vayer2019optimal}: for two graphs $\{\bm{C}_s, \bm{\mu}_s, \bm{F}_s\}$ and $\{\bm{C}_t, \bm{\mu}_t, \bm{F}_t\}$,  their FGW discrepancy is 
\begin{eqnarray}\label{eq:fgwf}
\begin{aligned}
&d_{fgw}(\{\bm{C}_s, \bm{F}_s\}, \{\bm{C}_t, \bm{F}_t\})\\
&=\sideset{}{_{\bm{T}\in\Pi(\bm{\mu}_s,\bm{\mu}_t)}}\min (\langle \bm{C}_{st} - 2\bm{C}_s\bm{T}\bm{C}_t^{T} + \bm{D}, \bm{T} \rangle)^{\frac{1}{2}},
\end{aligned}
\end{eqnarray}
where $\bm{D}=(\bm{F}_s\odot\bm{F}_s)\bm{1}_D\bm{1}_{N_t}^{\top} + \bm{1}_{N_s}\bm{1}_D^{\top}(\bm{F}_t\odot\bm{F}_t)^{\top} - 2\bm{F}_s\bm{F}_t^{\top}$ is the Euclidean distance matrix computed based on features. 
We can approximate the FGW discrepancy by our GWD modules as well --- just replace the line 1 in Algorithm~\ref{alg:ot-ppa} (or Algorithm~\ref{alg:ot-badmm}) with $\bm{C}_{st}=(\bm{C}_s\odot\bm{C}_s)\bm{1}\bm{\mu}_t^{\top}+\bm{\mu}_s\bm{1}^{\top}(\bm{C}_t\odot\bm{C}_t)^{\top}+\bm{D}$. 
Similar to (\ref{eq:approx}), the loss function becomes
\begin{eqnarray}
\begin{aligned}
&d_{fgw}^{2(M)}(\{\bm{B}^{(L)},\bm{F}_{b}^{(L)}\}, \{\bm{C}_i,\bm{F}_i\})\\
&=\langle \bm{C}_{st}-2\bm{B}^{(L)}\bm{T}^{(M)}\bm{C}_i^{\top} + \bm{D}^{(L)},~\bm{T}^{(M)} \rangle,
\end{aligned}
\end{eqnarray}
where the new terms are
\begin{eqnarray}
\begin{aligned}
\bm{D}^{(L)}=&(\bm{F}_b^{(L)}\odot\bm{F}_b^{(L)})\bm{1}_D\bm{1}_{N_i}^{\top} + \bm{1}_{N_i}\bm{1}_D^{\top}(\bm{F}_i\odot\bm{F}_i)^{\top}\\
&- 2\bm{F}_b^{(L)}\bm{F}_i^{\top},\\
\bm{F}_b^{(L)}=&\frac{1}{\bm{\mu}_b\bm{1}_D^{\top}} \sideset{}{_{k=1}^{K}}\sum g_k(\bm{z}_i) \bm{T}_k^{(L)}\bm{H}_k.
\end{aligned}
\end{eqnarray}

\section{Related Work}
\subsection{GW discrepancy and its applications}
As a pseudometric of structural data like graphs, GW discrepancy~\cite{chowdhury2018gromov} has been applied to many problems, $e.g.$, registering 3D point clouds~\cite{memoli2011gromov}, aligning protein networks of different species~\cite{xu2019scalable}, and matching vocabulary sets of different languages~\cite{alvarez2018gromov}. 
For the graphs with node attributes, the work in~\cite{vayer2019optimal} proposes FGW discrepancy, which combines the GW discrepancy between graph structures with the Wasserstein discrepancy~\cite{villani2008optimal} between node attributes. 
GW barycenters are proposed in~\cite{peyre2016gromov}, achieving the interpolation of multiple graphs. 
Recently, GW discrepancy is applied as objective functions when learning machine learning models. 
The work in~\cite{bunne2018} trains coupled generative models in incomparable spaces by minimizing the GW discrepancy between their samples. 
The work in~\cite{xu2019gromov} learns node embeddings for unaligned pairwise graphs based on their GW discrepancy. 
Most of the existing works calculate GW discrepancy by Sinkhorn iterations\cite{sinkhorn1967concerning}, whose complexity per iteration is $\mathcal{O}(N^3)$ for the graphs with $N$ nodes. 
The high computational complexity limits the applications of GW discrepancy. 
These years, many variants of GW discrepancy have been proposed, $e.g.$, the recursive GW discrepancy~\cite{xu2019scalable}, and the sliced GW discrepancy~\cite{vayer2019sliced}.
Although these works have achieved encouraging results in many tasks, none of them consider building Gromov-Wasserstein factorization models as we do.

\subsection{Graph clustering methods}
Graph clustering is significant for many practical applications, $e.g.$, molecules modeling~\cite{borgwardt2005protein} and social network analysis~\cite{yanardag2015deep}. 
Different from graph partitioning, which finds clusters of nodes in a graph, graph clustering aims at finding clusters for different graphs. 
The key to this problem is embedding unaligned graphs. 
Many methods have been proposed to attack this problem, and most of them can be categorized into kernel-based methods, $e.g.$, the Weisfeiler-Lehman kernel in~\cite{vishwanathan2010graph}.
In principle, these methods iteratively aggregate node features according to the topology of the graphs. 
Recently, such a strategy becomes learnable with the help of graph convolutional networks (GCNs)~\cite{kipf2016semi}. 
Many GCN-based methods have been proposed to embed graphs, $e.g.$, the large-scale embedding method in~\cite{nie2017unsupervised}, and the hierarchical embedding method in~\cite{ying2018hierarchical}.
However, these methods rely on the labels and the attributes of nodes, which are often inapplicable for unsupervised learning. 
Moreover, the embeddings achieved by them are often with low interpretability because of their deep and highly-nonlinear processes. 
Besides the methods above, the GW discrepancy-based methods in~\cite{peyre2016gromov,vayer2019optimal} provide another strategy --- learning a distance matrix for graphs based on their pairwise (fused) GW discrepancy and applying spectral clustering accordingly. 
This strategy is feasible even if the side information of nodes is not available, but its computational complexity is very high.  
Compared with the methods above, our GWF model provides another strategy to embed graphs in a scalable and theoretically-supportive way and make the embeddings interpretable in the framework of factorization models.

\section{Experiments}
To demonstrate the usefulness of our GWF model, we test it on four graph datasets and compare it with state-of-the-art methods on graph clustering. 
When implementing our GWF model, we set its hyperparameters as follows: the number of atoms is $K=30$; the number of layers in the GWD module is $M=50$; the weight of regularizer $\gamma$ is $0.01$ for the PPA-based GWD module and $1$ for the BADMM-based GWD module; the number of layers in the GWB module is $L=2$. 
For the graph data without node attributes, the parameters of our GWF model involve $\bm{V}_{1:K}$ and $\bm{z}_{1:I}$. 
For the graph data with node attributes, we will further learn node embeddings of the atoms, $i.e.$, $\bm{H}_{1:K}$. 
We use Adam~\cite{kingma2014adam} to train our GWF model. 
The learning rate of our algorithm is $0.05$, and the number of epochs is $10$. 
To accelerate the convergence of our algorithm, we apply a warm-start strategy: we randomly select $K$ observed graphs as initial atoms. 
For each atom, the number of its nodes is equal to that of the selected graph.
The code is at \url{https://github.com/HongtengXu/Relational-Factorization-Model}.

\begin{figure*}[t]
    \centering
    \subfigure[AIDS]{
    \includegraphics[height=3cm]{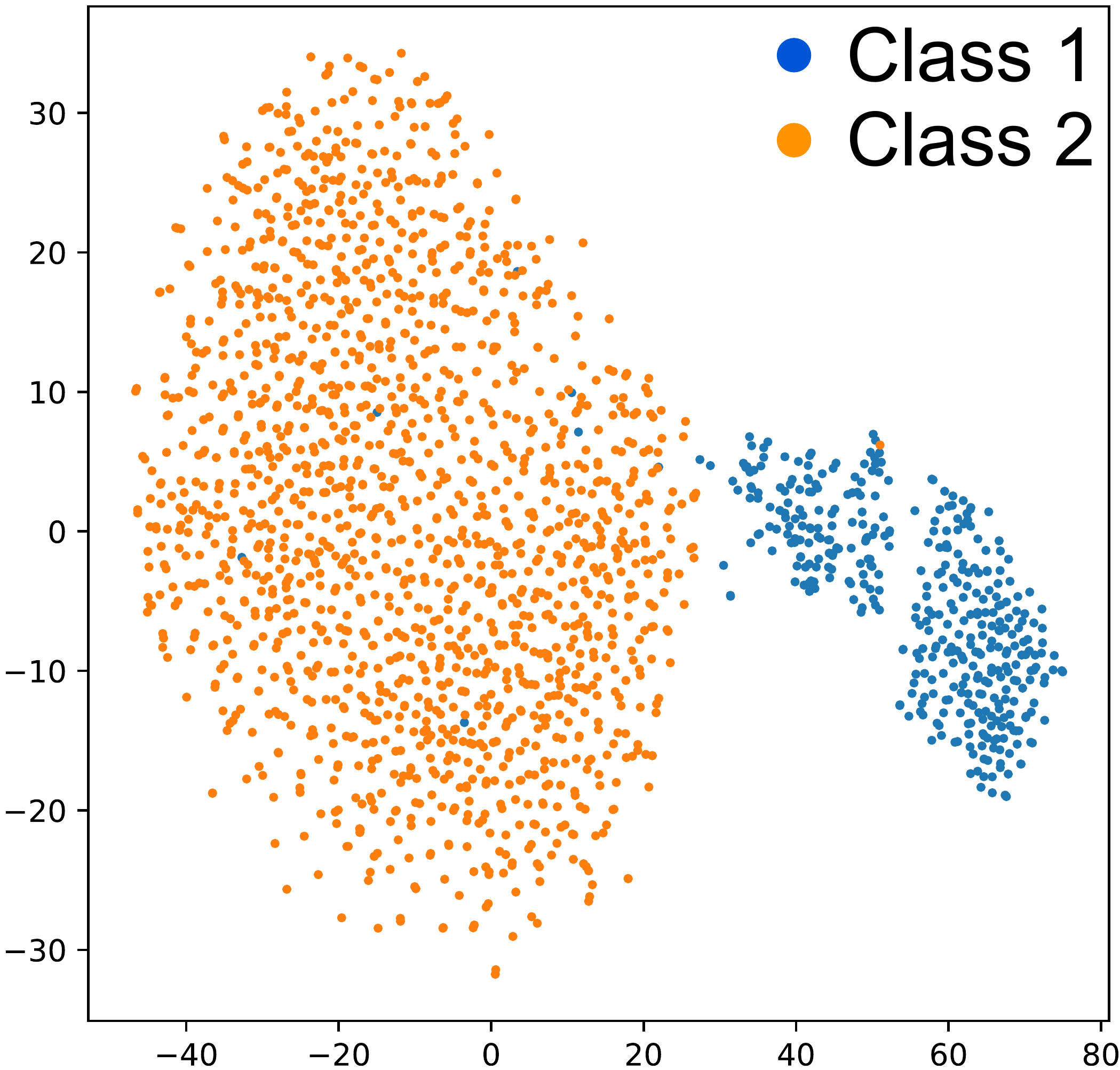}\label{fig:tsne_aids}
    }
    \subfigure[PROTEIN]{
    \includegraphics[height=3cm]{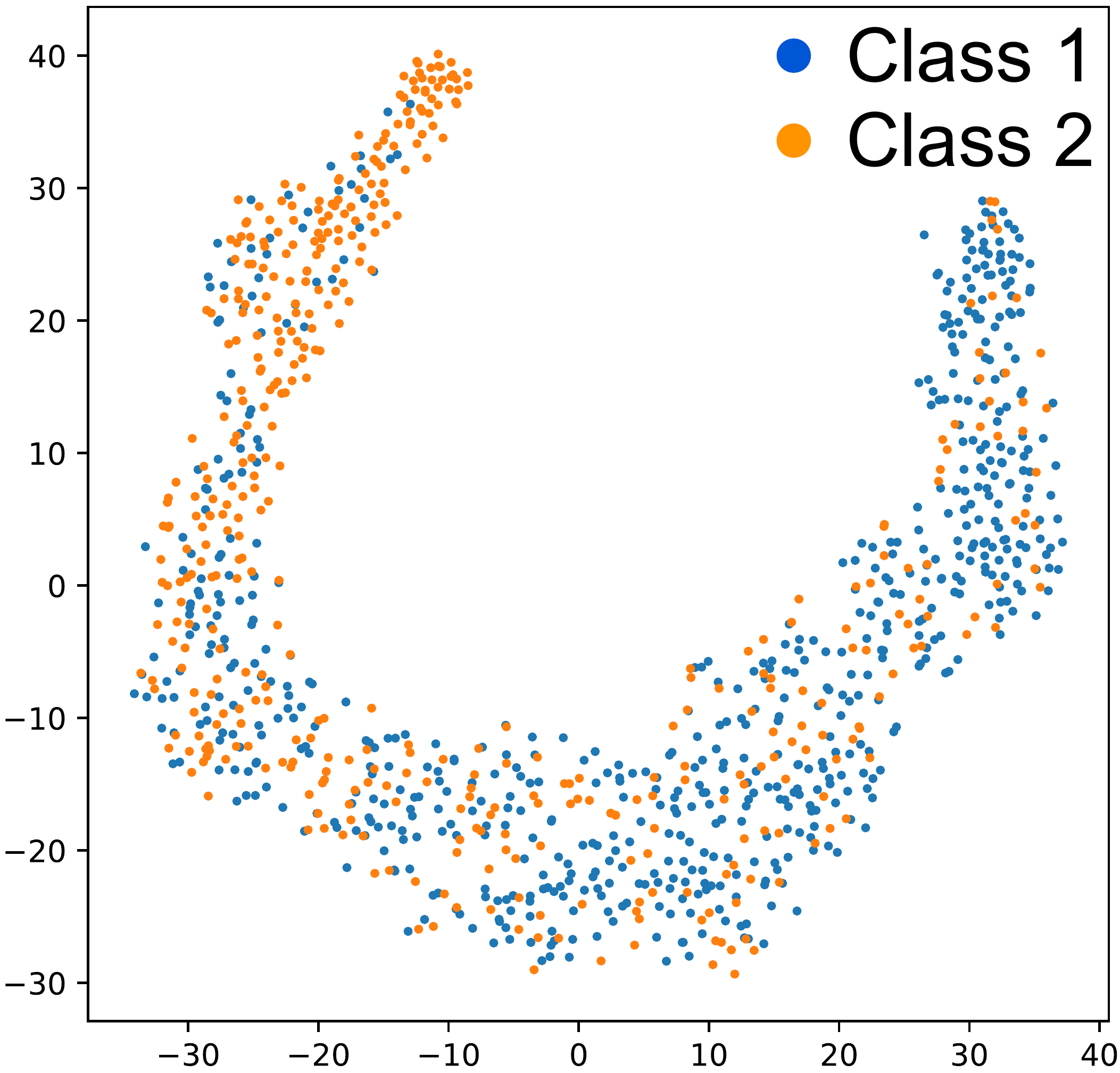}\label{fig:tsne_protein}
    }
    \subfigure[PROTEIN-F]{
    \includegraphics[height=3cm]{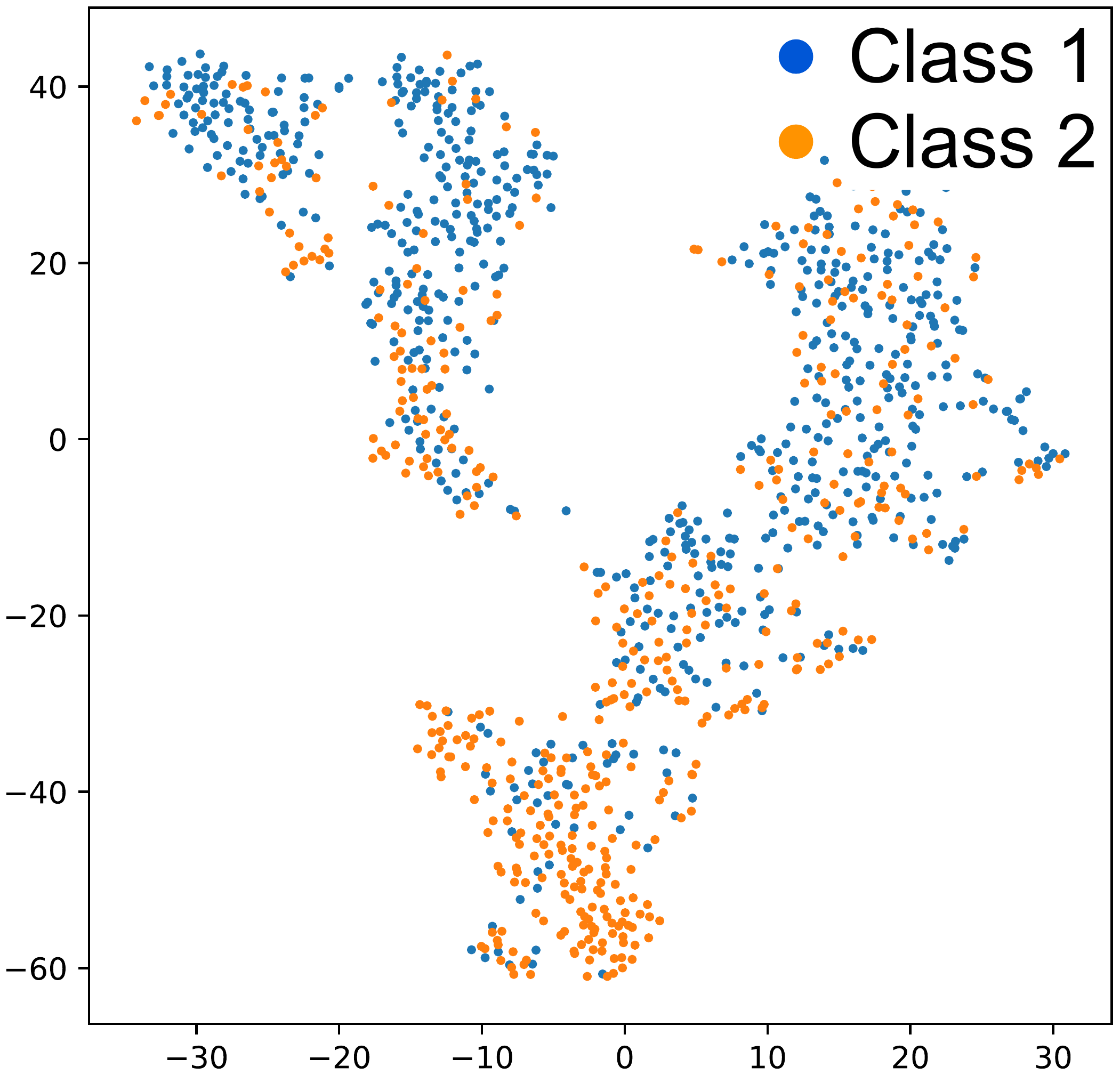}\label{fig:tsne_protein2}
    }
    \subfigure[IMDB-B]{
    \includegraphics[height=3cm]{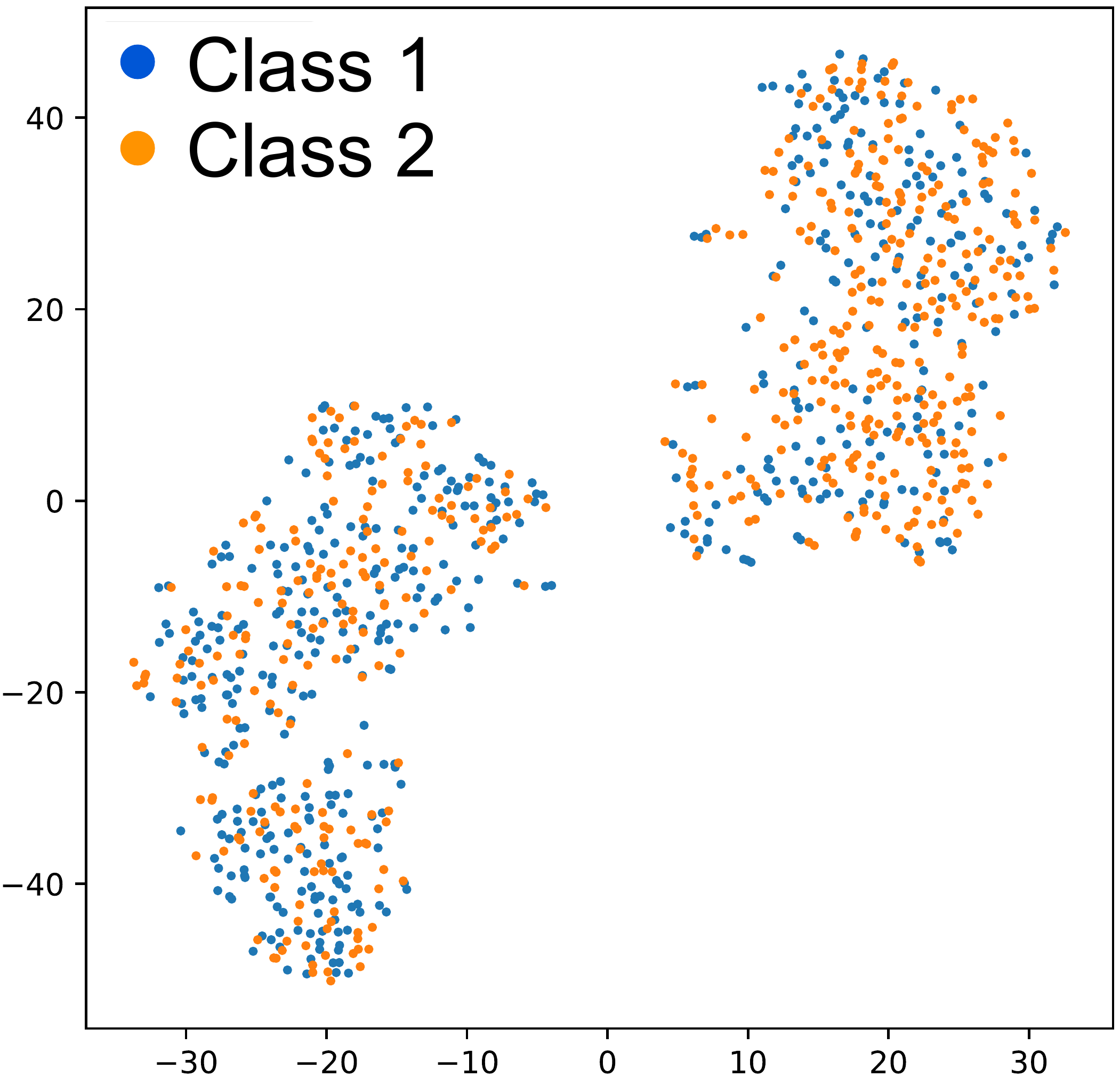}\label{fig:tsne_imdb}
    }
    \subfigure[Loss v.s. Epochs]{
    \includegraphics[height=3cm]{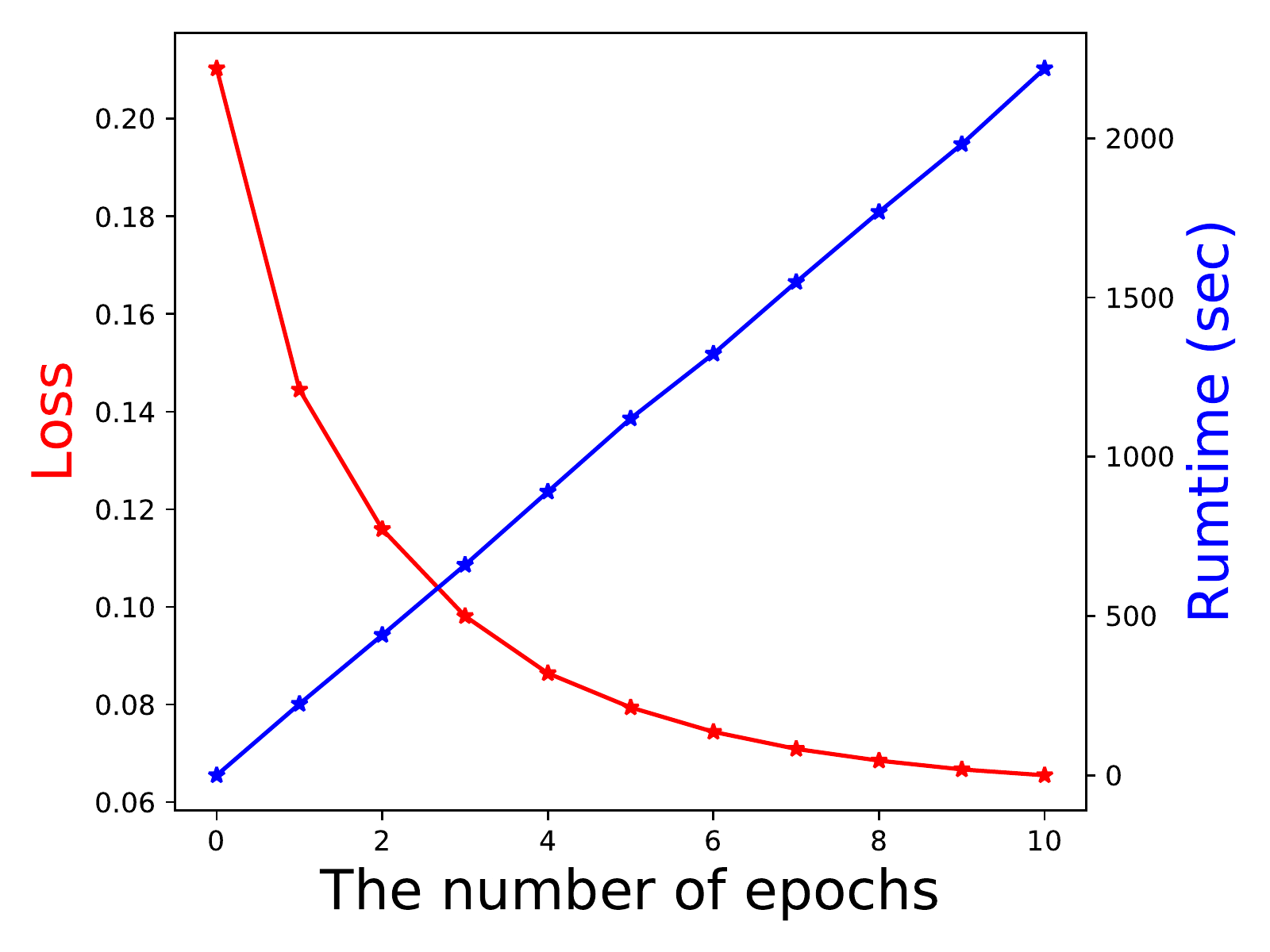}\label{fig:converge}
    }
    \caption{
    \small{(a-d) Visualizations of $\bm{z}_{1:I}$ based on t-SNE. (e) The convergence and the runtime of our method on IMDB-B.}
    }
    \label{fig:exp}
\end{figure*}

\subsection{Graph clustering}
We consider four commonly-used graph datasets~\cite{KKMMN2016} in our experiments: the AIDS~\cite{riesen2008iam}, the PROTEIN dataset and its full version PROTEIN-F~\cite{borgwardt2005protein}, and the IMDB-B~\cite{yanardag2015deep}. 
For each dataset, their graphs are categorized into two classes. 
The graphs in AIDS, PROTEIN, and PROTEIN-F represent molecules or proteins, which are with node attributes. 
The dimensions of the node attributes are $4$ for AIDS, $1$ for PROTEIN, and $29$ for PROTEIN-F. 
The graphs in IMDB-B represent user interactions in social networks, which merely contain topological information.
We show the details of the datasets in Table~\ref{tab:cmp}.

We apply various methods to cluster the graphs in each dataset and compare their performance in the aspect of clustering accuracy. 
For each dataset, we train a GWF model and applying K-means to learned embeddings $\bm{z}_{1:I}$.
Besides our GWF model, we consider two state-of-the-art methods as our baselines. 
The first is the fused Gromov-Wasserstein kernel method (FGWK) in~\cite{vayer2019optimal}. 
Given $N$ graphs, the FGWK method computes their pairwise fused GW discrepancy and construct a $N\times N$ kernel matrix $\exp(-d_{fgw} /\beta)$.\footnote{In~\cite{vayer2019optimal}, the authors claim that $\exp(-d_{fgw} /\beta)$ is a noisy observation of a true positive semidefinite kernel. Our implementation confirms its performance, but whether it can always be positive semidefinite may be questionable.}
Applying spectral clustering to the kernel matrix, we can cluster observed graphs into two clusters. 
When the node attributes are unavailable for the observed graphs, this method computes pairwise GW discrepancy instead to construct the kernel matrix. 
This method has been proven to be superior to traditional kernel-based methods on many datasets, including the PROTEIN and the IMDB-B datasets used in this work. 
The second competitor of our method is the K-means of graphs based on GW barycenter~\cite{peyre2016gromov} (GWB-KM). 
This method applies K-means to find centers of clusters iteratively. 
The centers are initialized randomly as two observed graphs. 
In each iteration, we first categorize the graphs into different clusters according to their GW discrepancy to the centers, and then we recalculate each center as the GW barycenter of the graphs in the corresponding cluster. 

Table~\ref{tab:cmp} shows that our GWF model outperforms its competitors consistently, which demonstrates its superiority in the task of graph clustering.\footnote{For the data with two clusters, given the ground truth labels $\bm{y}\in \{0,1\}^N$ and the estimated labels $\hat{\bm{y}}\in\{0, 1\}^N$, we calculate the clustering accuracy via $1-\frac{1}{N}\min(\|\bm{y}-\hat{\bm{y}}\|_1, \|\bm{y}-\bm{1}+\hat{\bm{y}}\|_1)$.} 
We implement our GWF model based on PPA and BADMM, respectively. 
The performance of the PPA-based model is better than that of the BADMM-based model in general because $i$) the graphs in the four datasets are undirected; $ii$) the BADMM-based model generally requires more steps to converge in the training phase, and $50$ steps may be too few for it.\footnote{When applying BADMM with $M=300$, its performance becomes close to that of PPA while the runtime is much longer. 
Unless the graphs are directed and the dataset is small, we prefer using the PPA-based method.}
Additionally, given $I$ graphs, FGWK calculates $\mathcal{O}(I^2)$ GW discrepancy, while both GWB-KM and our GWF model only need to calculate $\mathcal{O}(LKI)$ GW discrepancy. 
Because we can set $LK\ll I$, our model is more suitable for large-scale clustering problems. 
Figure~\ref{fig:exp} further visualizes the embeddings $\bm{z}_{1:I}$ for the four datasets based on t-SNE~\cite{maaten2008visualizing}.
The visualization results further verify the effectiveness of our GWF model --- for each dataset, the embeddings derived by our GWF model indeed reflect the clustering structure of the graphs. 
We show the convergence and the runtime of our learning method in Figure~\ref{fig:converge}.

\begin{table}[t]
	\centering
	\caption{Comparisons on clustering accuracy (\%)}\label{tab:cmp}
	\begin{small}
	\begin{tabular}{@{\hspace{0pt}}l@{\hspace{2pt}}|
	@{\hspace{2pt}}c@{\hspace{2pt}}|
	@{\hspace{2pt}}c@{\hspace{3pt}}c@{\hspace{2pt}}c@{\hspace{2pt}}c@{\hspace{0pt}}}
	\hline\hline
	    Dataset &\multirow{4}{*}{\# GWD}
	    &AIDS
	    &PROTEIN
	    &PROTEIN-F
	    &IMDB-B\\
	    \# graphs &
	    &2000
	    &1113
	    &1113
	    &1000\\
	    Ave. \#nodes&
	    &15.69
	    &39.06
	    &39.06
	    &19.77\\
	    Ave. \#edges&
	    &16.20
	    &72.82
	    &72.82
	    &96.53\\
		\hline
		FGWK  &$\mathcal{O}(I^2)$   
		&91.0$\pm$0.7  
		&66.4$\pm$0.8  
		&66.0$\pm$0.9   
		&56.7$\pm$1.5\\
		GWB-KM   &$\mathcal{O}(LKI)$            
		&95.2$\pm$0.9               
		&64.7$\pm$1.1               
		&62.9$\pm$1.3   
		&53.5$\pm$2.3\\
		GWF$_{\text{BADMM}}$ &$\mathcal{O}(LKI)$     
		&97.6$\pm$0.8               
		&69.2$\pm$1.0               
		&68.1$\pm$1.1   
		&55.9$\pm$1.8\\
		GWF$_{\text{PPA}}$   &$\mathcal{O}(LKI)$             
		&\textbf{99.5}$\pm$0.4 
		&\textbf{70.7}$\pm$0.7 
		&\textbf{69.3}$\pm$0.8 
		&\textbf{60.2}$\pm$1.6\\
	\hline\hline
	\end{tabular}
    \end{small}
\end{table}

Besides graph clustering, we also consider graph classification given the labels of graphs. 
In such a situation, we apply the learning strategy in (\ref{eq:ssl}) to learn our GWF model. 
The baselines include the kernel-based graph classification methods like the shortest path kernel (SPK)~\cite{borgwardt2005protein}, the HOPPER kernel (HOPPERK)~\cite{feragen2013scalable}, the propagation kernel (PROPAK)~\cite{neumann2016propagation}, the graphlet count kernel (GCK)~\cite{shervashidze2009efficient}, and the FGWK mentioned above. 
Each of these methods derives a kernel matrix and train a classifier based on kernel SVM. 
We test different methods based on 10-fold cross-validation. 
Table~\ref{tab:cmp2} shows the classification accuracy achieved by different methods on two datasets. 
Compared with the state-of-the-art method FGWK, our GWF model achieves comparable performance in the task of graph classification, and the fluctuations of our results are smaller than those of FGWK's results. 

\begin{table}[t]
	\centering
	\caption{Comparisons on classification accuracy (\%)}\label{tab:cmp2}
	\begin{small}
	\begin{tabular}{lc|lc}
	\hline\hline
	    Method
	    &PROTEIN
	    &Method
	    &IMDB-B\\
		\hline
		HOPPERK                   
		&71.6$\pm$3.7               
		&GCK   
		&56.9$\pm$4.0 \\
		PROPAK                     
		&60.3$\pm$5.1                
		&SPK   
		&56.2$\pm$3.1 \\
		FGWK                         
		&75.1$\pm$2.9   
		&FGWK                                   
		&64.2$\pm$3.3 \\
		GWF$_{\text{BADMM}}$    
		&71.4$\pm$3.6               
		&GWF$_{\text{BADMM}}$               
		&62.4$\pm$3.8 \\
		GWF$_{\text{PPA}}$           
		&73.7$\pm$2.0  
		&GWF$_{\text{PPA}}$                       
		&63.9$\pm$2.7 \\
	\hline\hline
	\end{tabular}
    \end{small}
\end{table}

\section{Conclusion and Future Work}
In this paper, we propose a novel Gromov-Wassserstein factorization model.
It is a pioneering work achieving an explicit factorization mechanism for graph clustering.
We design an efficient learning algorithm for learning this model with the help of  the envelope theorem. 
Experiments demonstrate that our model outperforms many existing methods in the tasks of graph clustering.
In the future, we plan to reduce the computational complexity of our learning algorithm further and consider its applications to large-scale graphs. 

\textbf{Acknowledgements}
This research was supported in part by DARPA, DOE, NIH, ONR, NSF, and Inifina ML. 

\bibliographystyle{aaai}
\bibliography{AAAI-XuH}

\begin{thebibliography}{}

\bibitem[\protect\citeauthoryear{Afriat}{1971}]{afriat1971theory}
Afriat, S.
\newblock 1971.
\newblock Theory of maxima and the method of lagrange.
\newblock {\em SIAM Journal on Applied Mathematics} 20(3):343--357.

\bibitem[\protect\citeauthoryear{Aharon, Elad, and
  Bruckstein}{2006}]{aharon2006k}
Aharon, M.; Elad, M.; and Bruckstein, A.
\newblock 2006.
\newblock K-svd: An algorithm for designing overcomplete dictionaries for
  sparse representation.
\newblock {\em IEEE Transactions on signal processing} 54(11):4311--4322.

\bibitem[\protect\citeauthoryear{Alvarez-Melis and
  Jaakkola}{2018}]{alvarez2018gromov}
Alvarez-Melis, D., and Jaakkola, T.~S.
\newblock 2018.
\newblock Gromov-{W}asserstein alignment of word embedding spaces.
\newblock In {\em EMNLP}.

\bibitem[\protect\citeauthoryear{Barab{\'a}si and
  others}{2016}]{barabasi2016network}
Barab{\'a}si, A.-L., et~al.
\newblock 2016.
\newblock {\em Network science}.
\newblock Cambridge university press.

\bibitem[\protect\citeauthoryear{Borgwardt \bgroup et al\mbox.\egroup
  }{2005}]{borgwardt2005protein}
Borgwardt, K.~M.; Ong, C.~S.; Sch{\"o}nauer, S.; Vishwanathan, S.; Smola,
  A.~J.; and Kriegel, H.-P.
\newblock 2005.
\newblock Protein function prediction via graph kernels.
\newblock {\em Bioinformatics} 21(suppl\_1):i47--i56.

\bibitem[\protect\citeauthoryear{Bunne \bgroup et al\mbox.\egroup
  }{2019}]{bunne2018}
Bunne, C.; Alvarez-Melis, D.; Krause, A.; and Jegelka, S.
\newblock 2019.
\newblock Learning generative models across incomparable spaces.
\newblock In {\em ICML}.

\bibitem[\protect\citeauthoryear{Cand{\`e}s \bgroup et al\mbox.\egroup
  }{2011}]{candes2011robust}
Cand{\`e}s, E.~J.; Li, X.; Ma, Y.; and Wright, J.
\newblock 2011.
\newblock Robust principal component analysis?
\newblock {\em Journal of the ACM} 58(3):11.

\bibitem[\protect\citeauthoryear{Chowdhury and
  M{\'e}moli}{2018}]{chowdhury2018gromov}
Chowdhury, S., and M{\'e}moli, F.
\newblock 2018.
\newblock The gromov-{W}asserstein distance between networks and stable network
  invariants.
\newblock {\em arXiv preprint arXiv:1808.04337}.

\bibitem[\protect\citeauthoryear{Feragen \bgroup et al\mbox.\egroup
  }{2013}]{feragen2013scalable}
Feragen, A.; Kasenburg, N.; Petersen, J.; de~Bruijne, M.; and Borgwardt, K.
\newblock 2013.
\newblock Scalable kernels for graphs with continuous attributes.
\newblock In {\em NeurIPS}.

\bibitem[\protect\citeauthoryear{Henaff, Bruna, and
  LeCun}{2015}]{henaff2015deep}
Henaff, M.; Bruna, J.; and LeCun, Y.
\newblock 2015.
\newblock Deep convolutional networks on graph-structured data.
\newblock {\em arXiv preprint arXiv:1506.05163}.

\bibitem[\protect\citeauthoryear{Kersting \bgroup et al\mbox.\egroup
  }{2016}]{KKMMN2016}
Kersting, K.; Kriege, N.~M.; Morris, C.; Mutzel, P.; and Neumann, M.
\newblock 2016.
\newblock Benchmark data sets for graph kernels.

\bibitem[\protect\citeauthoryear{Kingma and Ba}{2014}]{kingma2014adam}
Kingma, D.~P., and Ba, J.
\newblock 2014.
\newblock Adam: A method for stochastic optimization.
\newblock {\em arXiv preprint arXiv:1412.6980}.

\bibitem[\protect\citeauthoryear{Kipf and Welling}{2016}]{kipf2016semi}
Kipf, T.~N., and Welling, M.
\newblock 2016.
\newblock Semi-supervised classification with graph convolutional networks.
\newblock {\em arXiv preprint arXiv:1609.02907}.

\bibitem[\protect\citeauthoryear{Maaten and
  Hinton}{2008}]{maaten2008visualizing}
Maaten, L. v.~d., and Hinton, G.
\newblock 2008.
\newblock Visualizing data using t-sne.
\newblock {\em Journal of machine learning research} 9(Nov):2579--2605.

\bibitem[\protect\citeauthoryear{M{\'e}moli}{2011}]{memoli2011gromov}
M{\'e}moli, F.
\newblock 2011.
\newblock Gromov-{W}asserstein distances and the metric approach to object
  matching.
\newblock {\em Foundations of computational mathematics} 11(4):417--487.

\bibitem[\protect\citeauthoryear{Neumann \bgroup et al\mbox.\egroup
  }{2016}]{neumann2016propagation}
Neumann, M.; Garnett, R.; Bauckhage, C.; and Kersting, K.
\newblock 2016.
\newblock Propagation kernels: efficient graph kernels from propagated
  information.
\newblock {\em Machine Learning} 102(2):209--245.

\bibitem[\protect\citeauthoryear{Ng, Jordan, and Weiss}{2002}]{ng2002spectral}
Ng, A.~Y.; Jordan, M.~I.; and Weiss, Y.
\newblock 2002.
\newblock On spectral clustering: Analysis and an algorithm.
\newblock In {\em NeurIPS}.

\bibitem[\protect\citeauthoryear{Nie, Zhu, and Li}{2017}]{nie2017unsupervised}
Nie, F.; Zhu, W.; and Li, X.
\newblock 2017.
\newblock Unsupervised large graph embedding.
\newblock In {\em AAAI}.

\bibitem[\protect\citeauthoryear{Pearson}{1901}]{pearson1901liii}
Pearson, K.
\newblock 1901.
\newblock Liii. on lines and planes of closest fit to systems of points in
  space.
\newblock {\em The London, Edinburgh, and Dublin Philosophical Magazine and
  Journal of Science} 2(11):559--572.

\bibitem[\protect\citeauthoryear{Peyr{\'e}, Cuturi, and
  Solomon}{2016}]{peyre2016gromov}
Peyr{\'e}, G.; Cuturi, M.; and Solomon, J.
\newblock 2016.
\newblock Gromov-{W}asserstein averaging of kernel and distance matrices.
\newblock In {\em ICML}.

\bibitem[\protect\citeauthoryear{Riesen and Bunke}{2008}]{riesen2008iam}
Riesen, K., and Bunke, H.
\newblock 2008.
\newblock Iam graph database repository for graph based pattern recognition and
  machine learning.
\newblock In {\em Joint IAPR Workshop},  287--297.

\bibitem[\protect\citeauthoryear{Rolet, Cuturi, and
  Peyr{\'e}}{2016}]{rolet2016fast}
Rolet, A.; Cuturi, M.; and Peyr{\'e}, G.
\newblock 2016.
\newblock Fast dictionary learning with a smoothed {W}asserstein loss.
\newblock In {\em AISTATS}.

\bibitem[\protect\citeauthoryear{Schmitz \bgroup et al\mbox.\egroup
  }{2018}]{schmitz2018wasserstein}
Schmitz, M.~A.; Heitz, M.; Bonneel, N.; Ngole, F.; Coeurjolly, D.; Cuturi, M.;
  Peyr{\'e}, G.; and Starck, J.-L.
\newblock 2018.
\newblock Wasserstein dictionary learning: {O}ptimal transport-based
  unsupervised nonlinear dictionary learning.
\newblock {\em SIAM Journal on Imaging Sciences} 11(1):643--678.

\bibitem[\protect\citeauthoryear{Shervashidze \bgroup et al\mbox.\egroup
  }{2009}]{shervashidze2009efficient}
Shervashidze, N.; Vishwanathan, S.; Petri, T.; Mehlhorn, K.; and Borgwardt, K.
\newblock 2009.
\newblock Efficient graphlet kernels for large graph comparison.
\newblock In {\em AISTATS}.

\bibitem[\protect\citeauthoryear{Sinkhorn and
  Knopp}{1967}]{sinkhorn1967concerning}
Sinkhorn, R., and Knopp, P.
\newblock 1967.
\newblock Concerning nonnegative matrices and doubly stochastic matrices.
\newblock {\em Pacific Journal of Mathematics} 21(2):343--348.

\bibitem[\protect\citeauthoryear{Sra and Dhillon}{2006}]{sra2006generalized}
Sra, S., and Dhillon, I.~S.
\newblock 2006.
\newblock Generalized nonnegative matrix approximations with bregman
  divergences.
\newblock In {\em NeurIPS}.

\bibitem[\protect\citeauthoryear{Vayer \bgroup et al\mbox.\egroup
  }{2019a}]{vayer2019optimal}
Vayer, T.; Chapel, L.; Flamary, R.; Tavenard, R.; and Courty, N.
\newblock 2019a.
\newblock Optimal transport for structured data with application on graphs.
\newblock In {\em ICML}.

\bibitem[\protect\citeauthoryear{Vayer \bgroup et al\mbox.\egroup
  }{2019b}]{vayer2019sliced}
Vayer, T.; Flamary, R.; Tavenard, R.; Chapel, L.; and Courty, N.
\newblock 2019b.
\newblock Sliced gromov-{W}asserstein.
\newblock {\em arXiv preprint arXiv:1905.10124}.

\bibitem[\protect\citeauthoryear{Villani}{2008}]{villani2008optimal}
Villani, C.
\newblock 2008.
\newblock {\em Optimal transport: {O}ld and new}, volume 338.
\newblock Springer Science \& Business Media.

\bibitem[\protect\citeauthoryear{Vishwanathan \bgroup et al\mbox.\egroup
  }{2010}]{vishwanathan2010graph}
Vishwanathan, S. V.~N.; Schraudolph, N.~N.; Kondor, R.; and Borgwardt, K.~M.
\newblock 2010.
\newblock Graph kernels.
\newblock {\em Journal of Machine Learning Research} 11(Apr):1201--1242.

\bibitem[\protect\citeauthoryear{Wang and Banerjee}{2014}]{wang2014bregman}
Wang, H., and Banerjee, A.
\newblock 2014.
\newblock Bregman alternating direction method of multipliers.
\newblock In {\em NeurIPS}.

\bibitem[\protect\citeauthoryear{Wang and Blei}{2011}]{wang2011collaborative}
Wang, C., and Blei, D.~M.
\newblock 2011.
\newblock Collaborative topic modeling for recommending scientific articles.
\newblock In {\em KDD}.

\bibitem[\protect\citeauthoryear{Xu \bgroup et al\mbox.\egroup
  }{2019}]{xu2019gromov}
Xu, H.; Luo, D.; Zha, H.; and Carin, L.
\newblock 2019.
\newblock Gromov-{W}asserstein learning for graph matching and node embedding.
\newblock In {\em ICML}.

\bibitem[\protect\citeauthoryear{Xu, Luo, and Carin}{2019}]{xu2019scalable}
Xu, H.; Luo, D.; and Carin, L.
\newblock 2019.
\newblock Scalable {G}romov-{W}asserstein learning for graph partitioning and
  matching.
\newblock {\em arXiv preprint arXiv:1905.07645}.

\bibitem[\protect\citeauthoryear{Yanardag and
  Vishwanathan}{2015}]{yanardag2015deep}
Yanardag, P., and Vishwanathan, S.
\newblock 2015.
\newblock Deep graph kernels.
\newblock In {\em KDD}.

\bibitem[\protect\citeauthoryear{Ye \bgroup et al\mbox.\egroup
  }{2017}]{ye2017fast}
Ye, J.; Wu, P.; Wang, J.~Z.; and Li, J.
\newblock 2017.
\newblock Fast discrete distribution clustering using {W}asserstein barycenter
  with sparse support.
\newblock {\em IEEE Transactions on Signal Processing} 65(9):2317--2332.

\bibitem[\protect\citeauthoryear{Ying \bgroup et al\mbox.\egroup
  }{2018}]{ying2018hierarchical}
Ying, Z.; You, J.; Morris, C.; Ren, X.; Hamilton, W.; and Leskovec, J.
\newblock 2018.
\newblock Hierarchical graph representation learning with differentiable
  pooling.
\newblock In {\em NeurIPS}.

\end{thebibliography}

\end{document}